\documentclass[letterpaper, 10 pt, journal, twoside]{IEEEtran}

\usepackage{balance} 
\usepackage{amssymb}
\usepackage{graphics}
\usepackage{graphicx}
\usepackage{amsmath} % assumes amsmath package installed
\usepackage{floatrow}
\usepackage{multirow}
\usepackage{url}
\usepackage{svg}
\usepackage{textcomp}
\usepackage{gensymb}
\usepackage[ruled,vlined]{algorithm2e}
\usepackage[colorinlistoftodos]{todonotes}

\usepackage{multicol}

\RequirePackage{booktabs} % Required for better table rules
\usepackage{multirow}

\usepackage{caption}
\usepackage{subcaption}
\graphicspath{ {./Figures/} }

\usepackage{pifont}% http://ctan.org/pkg/pifont
\newcommand{\cmark}{\ding{51}}%
\newcommand{\xmark}{\ding{55}}%

\begin{document}

\title{Continual Learning of  Hand Gestures for Human-Robot Interaction}

\author{Xavier Cucurull$^{1}$ and Anaís Garrell$^{1}$ % <-this % stops a space

% \thanks{$^{1}$Xavier Cucurull, Alberto Sanfeliu and Anaís Garrell are with the Institut de Robòtica i Informàtica Industrial (CSIC-UPC). Llorens Artigas 4-6, 08028, Barcelona, Spain.
%         {\tt\small \{xavier.cucurull, alberto.sanfeliu, anais.garrell\}@upc.edu}}
% \thanks{Work supported under the Spanish State Research Agency through the ROCOTRANSP project (PID2019-106702RB-C21 / AEI / 10.13039/501100011033)) and the EU project CANOPIES (H2020- ICT-2020-2-101016906)}
%\thanks{Digital Object Identifier (DOI): see top of this page.}
}

%\markboth{IEEE Robotics and Automation Letters. Preprint Version. Accepted July, 2022}
%{Peral \MakeLowercase{\textit{et al.}}: Efficient Hand Gesture Recognition for HRI} 

\maketitle
% ------------------------------------------------
\begin{abstract} % 120
% compare to icarl jetcas (event-based, hundred of instances)
In this paper, we present an efficient method to incrementally learn to classify static hand gestures. This method allows users to teach a robot to recognize new symbols in an incremental manner. Contrary to other works which use special sensors or external devices such as color or data gloves, our proposed approach makes use of a single RGB camera to perform static hand gesture recognition from 2D images. Furthermore, our system is able to incrementally learn up to 38 new symbols using only 5 samples for each old class, achieving a final average accuracy of over 90\%. In addition to that, the incremental training time can be reduced to a 10\% of the time required when using all data available.
% TODO: the time reduction was only observed on the kaggle asl dataset. maybe should also discuss kaggle asl results

% TODO: main contributions. New Dataset?
%\keywords{}
\end{abstract}

% ------------------------------------------------

\section{Introduction}\label{sec_introduction}

When we think about a natural form of communication among humans, the use of speech is one of the first things that comes to mind. However, human communication is multimodal, with speech and gestures coexisting and having both great importance \cite{tomasello2010origins, levinson2014origin}. In infants, global motor development occurs earlier than oral language and as a result, the first form of communication appears in the form of gestures \cite{vallotton2008signs}, such as finger pointing, and can be used as an effective form of early communication.

In the case of human-robot interaction (HRI), during the past decades we have evolved from simple teleoperation using remote controls, keyboards and joysticks, to a relationship that is increasingly becoming more natural and similar to the communication that takes place among humans. With the appearance of collaborative \cite{repiso2018robot} and assistive robots \cite{peshkin2001cobot}, using natural forms of interaction is becoming more and more important. This work focuses on the use of hand gesture recognition by making use of an RGB camera, to simplify the needed equipment, and to offer a cost-effective solution.

If hand gestures are used to communicate with a robot with the goal of providing a natural and easy way of interaction, it might be desirable to teach the robot to recognize new symbols. 
%In supervised machine learning, the traditional way of teaching a system consists of using a dataset of labeled data that is used to train the algorithm with. If a new symbol needs to be taught to the system, first, data from the new gesture needs to be gathered and added to the original dataset. Then, the model needs to be retrained from scratch on the new extended database. However, this might not be practical, as storing all the data used to train the original model might use a lot of memory and, most importantly, retraining a model from scratch might also take too much time. Unlike the human brain, artificial neural networks suffer from catastrophic forgetting \cite{McCloskey1989}, a phenomenon by which a model forgets its previous knowledge when trained on new data. 
Continual Learning (CL), which is a machine learning paradigm by which a system can incrementally learn new information without forgetting what it has previously learned \cite{Lesort2020}, can be used to %overcome the aforementioned problems, and to 
provide a system with a more natural way of learning. To this end, we propose an approach based on deep learning, called HAGIL (HAnd Gesture Incremental Learning), a system that can incrementally learn to classify static hand gestures from RGB images and that is targeted at human robot interaction.

\begin{figure}[!tb]
\centering
    \includegraphics[width=1\textwidth]{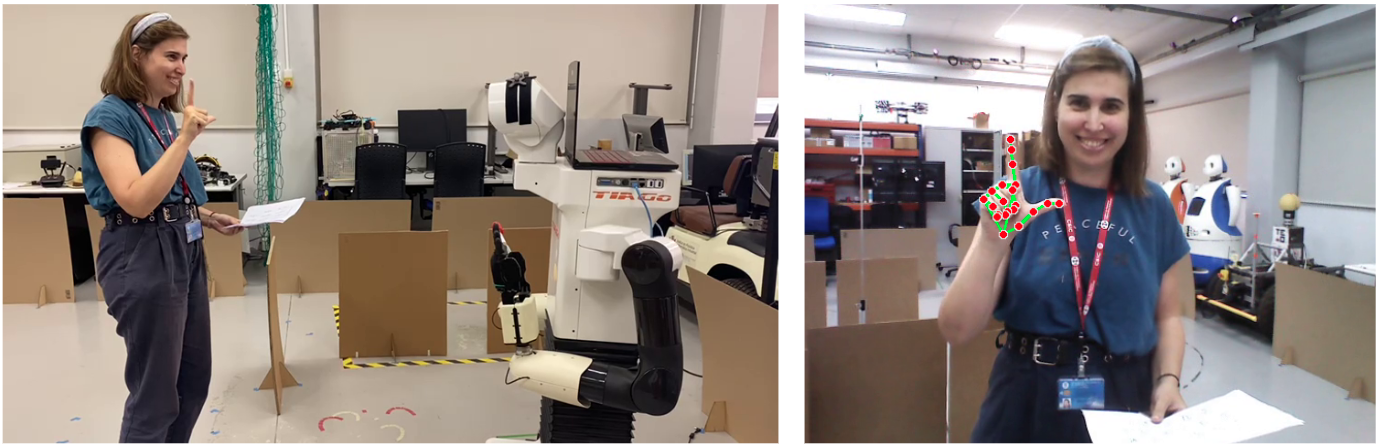}
    \caption[Volunter acting in front of the Tiago robot]{The proposed approach is used to teach a robot to incrementally recognize new hand gestures.}
    \label{fig:anais_tiago_example}
\end{figure}

\begin{figure*}[!th]
    \centering
    \includegraphics[width=\textwidth]{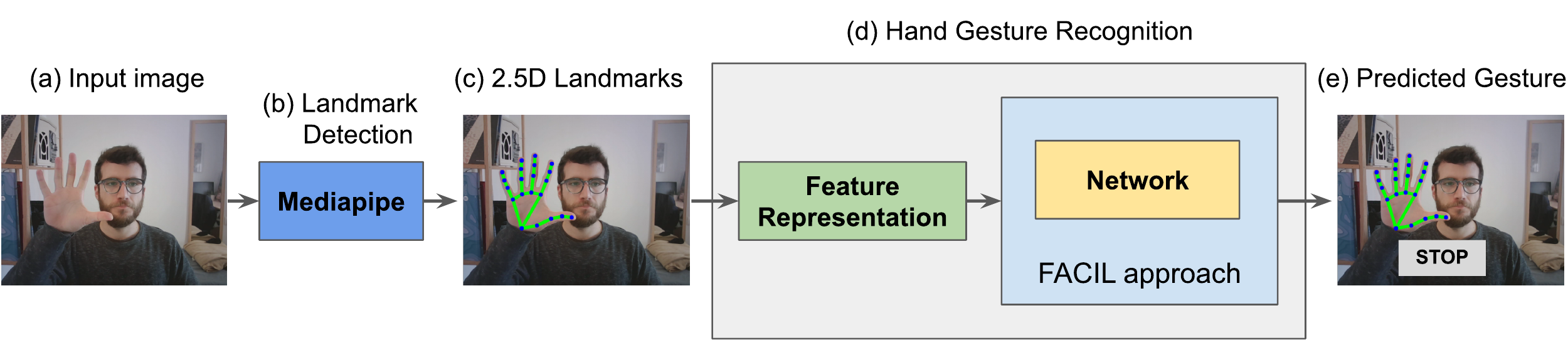}
    \caption[HAGIL System Overview]{Overview of the proposed approach for hand gesture incremental learning. Given an input image (a), the proposed approach uses the MediPipe Hands pose detector (b) to extract the landmarks (c) of the hands present in the input. Next, the detected landmarks are encoded into a feature vector representation, which is then used to perform hand gesture recognition through the use of a NN (d). To incorporate CL capabilities, the NN classifier is encapsulated into a modified version of the FACIL \cite{masana2020class} incremental learning framework.}
    \label{fig:hagil_system_overview}
\end{figure*}

An overview of this approach is shown in Fig. \ref{fig:hagil_system_overview}. The proposed method can be divided into two modules. The first one is dedicated to the extraction of visual features, more particularly hand landmarks (joints), from the input image. In this work, we rely on the open-source Mediapipe framework \cite{vakunov2020mediapipe}, which offers a lightweight and reliable hand landmark detector. The second module encodes the extracted landmarks into a feature representation, which is fed into a neural network (NN) in order to predict static hand gestures. In order to implement continual learning capabilities, the neural network classifier is encapsulated into a modified version of the FACIL \cite{masana2020class} incremental learning framework, which implements various incremental learning (IL) approaches. 

% ------------------------------------------------
% ------------------------------------------------

Finally, our method is thoroughly evaluated on a novel static hand gestures dataset that combines the existing UniPd datasets \cite{marin2014hand, memo2015exploiting, marin2016hand, memo2018head} with a custom-made dataset, accounting for a total of 38 unique gestures performed by different subjects. In addition, two American Sign Language (ASL) alphabet datasets are also used to evaluate the performance of the proposed HAGIL system.

% TODO: asl datasets
% which will then be applied and evaluated on the ASL Finger Spelling and ASL Alphabet datasets.
%ASL Finger Spelling \cite{pugeault2011spelling}            
%Kaggle ASL Alphabet \cite{nagaraj2018ASL}  \cite{rasband2018ASL}

The contribution of this paper is threefold: first, a deep learning approach for continual learning of static hand gestures called HAGIL is proposed. Second, we evaluate and study different incremental learning strategies to determine which components contribute the most to avoiding catastrophic forgetting. An extensive evaluation of the IL system is performed under a variety of parameters and datasets, to find a balance between accuracy and resource consumption, and to demonstrate its applicability and efficacy. Finally, this work introduces a new dataset that consists of 21 hand gestures that are performed by 4 distinct subjects, and which can be used as a benchmark in hand gesture continual learning.

The remainder of this paper is organized as follows. Section II presents the related work of hand gesture recognition, and continual learning. Section III describes the components of the proposed method.%: hand landmarks detection, static hand gesture recognition, and incremental learning of symbols. 
In Section IV, the method is evaluated in a state-of-the-art dataset. Finally, some conclusions and future work are provided in Section V.

% ------------------------------------------------

\section{Related Work}\label{sec_relatedwork}
In this section, we present some related work in the fields of %human-robot interaction,  
static hand gesture recognition, and continual learning methods applied in the field of robotics.

%\subsection{Human-Robot Interaction}\label{subsec_hri}
%Although robot technology was primarily developed in the mid and late 20th century, the multidisciplinary field of Human-Robot Interaction (HRI) began to gain importance in the mid 1990s \cite{goodrich2008human}. In recent years, the emergence and proliferation of assistive and collaborative robots has increased the need to consider a more social dimension of HRI, and thus more natural ways of interaction are being researched \cite{repiso2020people}. 

%As in communication among humans, robots and humans can exchange information using the senses of hearing, seeing, and touching. In HRI, these types of communication can take different forms, which include the use of mobile devices \cite{keskinpala2003pda}, speech based communication \cite{atrash2009development}, and gestures including hand and facial movements \cite{xiao2014human}. Since human communication is considered to be multimodal, many authors have investigated ways to incorporate multimodal communication into human-robot interaction \cite{perzanowski2001building} \cite{stiefelhagen2007enabling}.

\subsection{Hand Gesture Recognition}\label{subsec_rel_handgest_rec}

As mentioned above, the use of hand gestures constitutes a natural form of communication among humans and can therefore be an effective method for natural and accessible HRI. In addition to that, hand gestures might offer a better alternative to speech recognition to overcome challenges such as noise, reverberation and distant speech \cite{ravanelli2020multi,laplaza2022body}. As a result, hand gesture recognition has been thoroughly studied in the field of HRI.

In the late 1980s and 1990s, several authors proposed the use of data gloves equipped with sensors \cite{zimmerman1986hand, liang1995real} to recognize hand gestures. Later, \cite{iwai1996gesture} introduced the use of colored gloves to detect hand regions. Following the appearance of the Microsoft Kinect, many authors proposed the use of depth sensors and RGB-D images \cite{zafrulla2011american, tao2018american}. Other sensor-based approaches include the use of radar \cite{molchanov2015multi} and reflected impulses \cite{kim2017hand}. Regarding hand gesture recognition from RGB images, the extraction of features was initially performed using techniques such as histogram of oriented gradients (HOG) \cite{freeman1995orientation, feng2013static, prasuhn2014hog} and later by applying CNNs \cite{ahuja2019convolutional, islam2019static, mazhar2019real}.

Lately, pose estimation algorithms such as OpenPose \cite{cao2017realtime}, Mediapipe Holistic \cite{vakunov2020mediapipe} and OpenMMLab MMpose \cite{mmpose2020} have allowed skeleton-based hand gesture recognition \cite{de2016skeleton, moryossef2021evaluating, peral2022efficient} from RGB images. With respect to the classification methods employed, the most common algorithms used in the literature are SVM \cite{marin2014hand, memo2015exploiting}, kNN \cite{chuan2014american, aryanie2015american} and CNN \cite{garcia2016real, abdulhussein2020hand}.

%Although many works have been proposed in the field of hand gesture recognition, comparing the accuracy performance of the different methods is a complicated task, as there is not a common benchmark dataset that can be used to assess the quality of the different techniques. Despite the fact that various authors have reported an accuracy of around 99\% in the classification of American Sign Language fingerspelling \cite{aryanie2015american, abdulhussein2020hand}, most of the times these results cannot be translated into a real-world scenario because of the applied methodology. For instance, \cite{ranga2018american} and \cite{barbhuiya2021cnn} show that using data samples from different subjects in the train and test splits compared to mixing them, can reduce the performance of state-of-the-art algorithms from 95\% to 50\%. For that reason, the procedure used to create the training, validation, and test splits is crucial to build a system that can perform with good results when deployed and used in real life.

Most of the presented approaches require external devices such as sensors, depth cameras or color gloves. Our proposal uses a skeleton-based approach with just an RGB camera. This approach allows the user to interact with a robot in a natural way, without being required to wear additional equipment, in a cost-effective manner.

\subsection{Continual Learning in Robotics}\label{subsec_cl_robotics}
In continual learning, the system must continuously learn new tasks without forgetting what it has previously learned. This type of learning is especially interesting for robotics, where an autonomous agent might need to constantly learn from its interaction with the environment \cite{Lesort2020}. Below, we review various examples of continual learning with neural networks applied to visual tasks in robotics.

In \cite{fanello2013icub}, the \textit{Hello iCubWorld} dataset, consisting of images of 7 objects captured with a humanoid robot, was introduced. Using the \textit{iCubWorld28} dataset, \cite{pasquale2015teaching} trained a humanoid robot to incrementally classify objects using a CNN trained on ImageNet \cite{russakovsky2015imagenet} as a fixed feature extractor and a Regularized Least Squares (RLS) \cite{rifkin2003regularized} classifier on top of it. Other works also make use of ImageNet pre-trained networks as fixed feature extractors and incrementally learn to classify images using nearest-centroid-based classifiers using t-SNE \cite{pasquale2016object} or Agg-Var \cite{ayub2020tell} clustering. To address the lack of benchmark datasets specifically designed for continuous object recognition, \cite{lomonaco2017core50} introduced \textit{CORe50}, a collection of 50 domestic objects captured in a sequence where the objects move slightly in front of the camera, and presented CWR (Copy Weights with Re-init), an architectural CL strategy based on a VGG \cite{simonyan2014very} CNN. In \cite{efthymiou2021visual}, a visual system with incremental learning for the detection of actions from child-robot interaction is
proposed. In their work, a Temporal Segment Network (TSN) \cite{wang2016temporal} is used coupled with iCaRL \cite{rebuffi2017icarl} for continual learning of actions.

Concerning continual learning for static hand gesture classification, \cite{lungu2019incremental} also used iCaRL to incrementally learn hand symbols captured by event-based cameras, learning up to 16 symbols are learned with a final classification accuracy of 80\%. In this paper, we propose a framework to learn a larger number of symbols with a higher classification accuracy by using RGB still images captured by a standard CMOS camera, a more cost-effective option compared to event-based systems. 

% ------------------------------------------------
% TODO: 
\section{Proposed Method}\label{sec_system_overview}

In this section, we describe the proposed method, HAGIL, which is depicted in Fig. \ref{fig:hagil_system_overview}. It consists of two main modules, one to detect hand landmarks, and the second one dedicated to perform continual learning of hand gestures by means of a neural network. In the following, these modules are explained in more detail.

\subsection{Hand Landmarks Detection}\label{subsec_landmark_detect}
As seen above, our project uses the open-source MediaPipe framework for real-time extraction of hand landmarks. The use of a skeleton-based approach, compared to using an image-based gesture recognition method, allows the system to be more robust to variations in terms of illumination, subject characteristics, and background. Consequently, a robot can be trained with more limited data in a room with artificial lighting and a clear background and eventually be used in a different setting such as outdoors, with natural lighting and a cluttered background, and still perform with great accuracy.

%\begin{figure}[!tb]
%\centering
%    \includegraphics[width=1\textwidth]{Figures/canopies.png}
%    \caption[Volunter acting in front of the Tiago robot]{The proposed approach is used .}
%    \label{fig:anais_tiago_example}
%\end{figure}

% ------------------------------------------------
% \begin{figure}[!tb]
%     \begin{subfigure}[b]{0.3\textwidth}
% 		\centering
% 		\includegraphics[width=\textwidth]{mp_example4.png}
% 	\end{subfigure}
%     \hfill
% 	\begin{subfigure}[b]{0.3\textwidth}
% 		\centering
% 		\includegraphics[width=\textwidth]{mp_example6.png}
% 	\end{subfigure}
% 	\hfill
% 	\begin{subfigure}[b]{0.3\textwidth}
% 		\centering
% 		\includegraphics[width=\textwidth]{mp_example3.png}
% 	\end{subfigure}
% 	\caption{Example images showing the landmarks detected by Mediapipe in various angles and orientations.}
% 	\label{fig:landmarks_examples}
% \end{figure}
% ------------------------------------------------

%MediaPipe's landmark detection pipeline consists of two models working together. First, a palm detector takes the input image and returns a bounding box for the hand. Then, a hand landmark model uses the cropped hand bounding box to extract 21 hand landmarks. In a continuous hand landmark extraction scenario, this two-stage approach reuses the bounding box detections of previous frames in order to speed up the process, allowing the system to perform in real-time on mobile devices.

\subsection{Hand Gesture Recognition}\label{subsec_handgest_rec}
The recognition of static hand gestures is performed by a neural network model that uses the hand landmarks provided by the MediaPipe detector. The input to this NN can simply be the raw landmark coordinates $(x, y, z)$ of the detected joints. However, in this section, we describe the feature engineering process used to select a robust feature representation that leads to a better classification performance, and we present the neural network architecture for hand gesture recognition. In these initial tests, the \textit{UniPd} dataset, which consists of 17 unique static hand gestures, is used for evaluation. For these experiments, only the training and validation splits are used, since the test split is kept apart for the final experiments.
 
\subsubsection{Feature Representation}

As seen in Figure \ref{fig:hagil_system_overview}, each hand consists of 21 landmarks. MediaPipe returns 2.5D landmarks (the term 3D is not used because the depth is relative to the wrist), resulting in 21 (x, y, z) coordinates that can be used to represent each hand pose. A naive feature representation can be obtained by flattening the coordinates into a $21 \times 3 = 63$ values array. Alternatively, if only the 2D coordinates are used, this results in a $21 \times 2 = 42$ values array. In our experiments, using only the 2D coordinates gave the best results, so the third axis is discarded. This first feature representations can be referred to as \textit{Raw}, and can work really well when the hands in the training and validation images are located in similar positions (e.g. centered on the frame). However, if the hand moves around, for example to a corner of the image, the values of the joint coordinates will drastically change and model performance will decrease. For that reason, a translation invariant feature representation shall be used in order to cope with this problem. 

The first method to achieve a translation invariant feature vector is to calculate the distance between each landmark and the wrist, which is the most invariable joint of the hand. Equation \ref{eq:wrist_distance} shows how the distance between a joint \textit{j} and the wrist \textit{w} is calculated.

\begin{equation} \label{eq:wrist_distance}
d(j, w) = (x_{j} - x_{w}, y_{j} - y_{w})
\end{equation}

This feature representation, referred to as \textit{WristDiff} results in a $(21 - 1) \times 2 = 40$ values array. The second method studied, called \textit{WristEuclidean2D}, computes the Euclidean distance from each landmark to the wrist, resulting in a feature vector of $21 - 1 = 20$ values. Another feature representation alternative that was explored is based on calculating the distances among each of the hand landmarks. In the first approach, \textit{AllDiff} computes the difference between each of the landmarks to each of the others, resulting in a feature vector of $(21^{2} - 21) \times 2 = 420$. Alternatively, \textit{AllEuclidean}, comptues the Euclidean distance within all landmarks, resulting in a feature vector of $21^{2} - 21 = 210$ values.

The classification accuracy for each feature representation method, averaged on 10 runs, can be seen in Table \ref{tab:feature_representation_acc}. These results display that the three best feature representations, according to the classification accuracy achieved, are \textit{AllDiff}, \textit{WristDiff}, and \textit{AllEuclidean}, with an average accuracy above 98.10\%. With the goal of obtaining a robust feature representation that is translation invariant but rotation dependent, the three best representations are combined, resulting in the best overall classification accuracy. After concatenating \textit{AllEuclidean + WristDiff + AllDiff}, a final feature vector of $210 + 40 + 420 = 670$ values is obtained.

\begin{table}[!ht]
    \centering
    \begin{tabular}{ll}
    \toprule
        \textbf{Features}                       & \textbf{Accuracy (\%)}    \\ \midrule
        Raw                                     & 94.4 ± 1.49               \\
        WristDiff                               & \textbf{98.27 ± 0.41}     \\
        WristEuclidean                          & 84.87 ± 0.62              \\
        AllEuclidean                            & \textbf{98.12 ± 0.49}     \\
        AllDiff                                 & \textbf{98.31 ± 0.46}     \\
        AllEuclidean+WristDiff+AllDiff          & \textbf{98.57 ± 0.48}     \\ \bottomrule
    \end{tabular}
    \caption[Classification accuracy for different feature representations]{Average classification accuracy in \% (± standard deviation) for different feature representations evaluated on the UniPd dataset.}
    \label{tab:feature_representation_acc}
\end{table}

\subsubsection{Network Architecture}

\begin{figure}[!tb]
    \centering
    \includegraphics[width=1\textwidth]{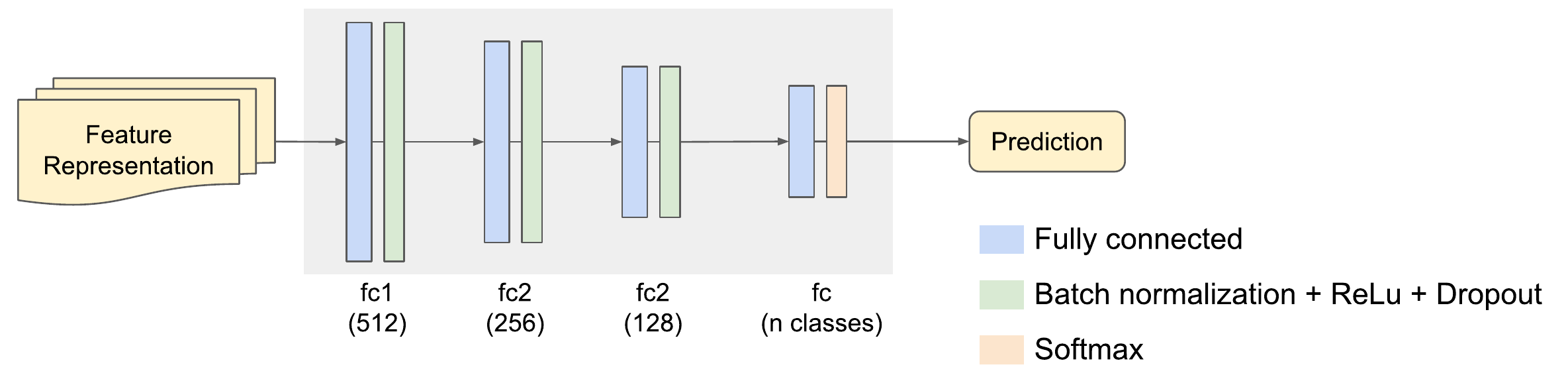}
    \caption{Layers diagram of the HagilNet network architecture.}
    \label{fig:hagilnet_diagram}
\end{figure}

After selecting the best feature representation, the next step was to choose the right neural network architecture. In this process, the \textit{UniPd} daset was used once again for evaluation. Initially, the width of the network (i.e. the number of units of the hidden layer) was increased by a factor of 2. Next, the length (i.e. the number of layers) was incremented successively by adding one layer. Adam \cite{kingma2014adam} was used as the optimizer. The initial learning rate was set to 0.001, and was decreased by a factor of 3 when the loss did not decrease for 5 consecutive epochs. In addition, various dropout probabilities were evaluated, obtaining the best results for (\textit{p}=0.35). Figure \ref{fig:hagilnet_diagram} depicts the chosen network architecture.

Finally, an ablation test was made to evaluate the contribution of batch normalization and dropout to the model accuracy. As observed in Table \ref{tab:ablation_study}, batch normalization is the regularization technique that contributes the most to the model performance.

\begin{table}[!ht]
\centering
    \begin{tabular}{@{}ccll@{}}
        \toprule
        \textbf{Dropout} & \textbf{Batch Normalization} & \textbf{Accuracy (\%)} \\ \midrule
        \cmark           & \cmark                       & 99.26 ± 0.26           \\
        \xmark           & \cmark                       & 99.02 ± 0.39           \\
        \xmark           & \xmark                       & 97.12 ± 0.89           \\ \bottomrule
        \end{tabular}
    \caption[Ablation study on the regularization methods applied]{Ablation study on the regularization methods applied. Average classification accuracy in \% (± standard deviation), evaluated on the UniPd dataset.}
    \label{tab:ablation_study}
\end{table}

% ------------------------------------------------
\subsection{Continual Learning Framework}\label{subsec_cl_framework}

In \cite{masana2020class}, the Framework for Analysis of Class-Incremental Learning (FACIL) was introduced. FACIL implements many state-of-the-art regularization and rehearsal incremental learning strategies focused on the task of class-incremental image recognition. In order to adapt the framework to the use of skeleton-based features, some modifications to the code were made. Since FACIL was designed to analyze and compare various incremental learning approaches, the entire dataset used for the evaluation should be provided at the beginning of the training phase. In order to be able to successively train the model with new data, HAGIL extends the existing framework by storing a copy of the model that can be subsequently re-trained incrementally. In addition to that, more changes were made to the original code to be able to use multiple datasets and different number of epochs for the initial and incremental tasks.

\subsection{Custom HandGest Dataset} \label{section:custom_datasets}
\begin{figure}[t]
    \begin{subfigure}[b]{0.24\textwidth}
		\centering
		\includegraphics[width=\textwidth]{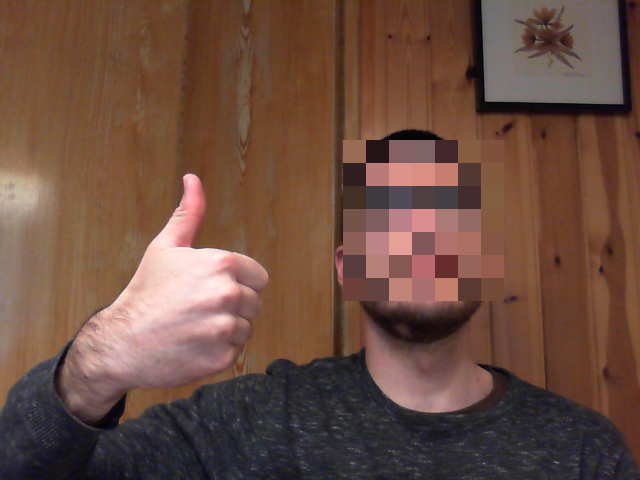}
	\end{subfigure}
    \hfill
	\begin{subfigure}[b]{0.24\textwidth}
		\centering
		\includegraphics[width=\textwidth]{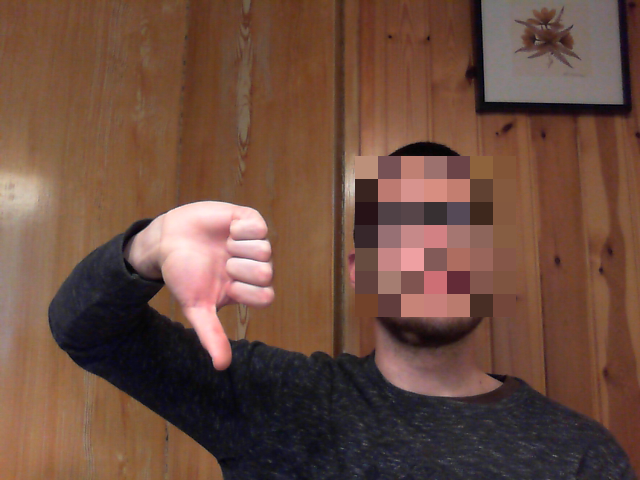}
	\end{subfigure}
    \hfill
	\begin{subfigure}[b]{0.24\textwidth}
		\centering
		\includegraphics[width=\textwidth]{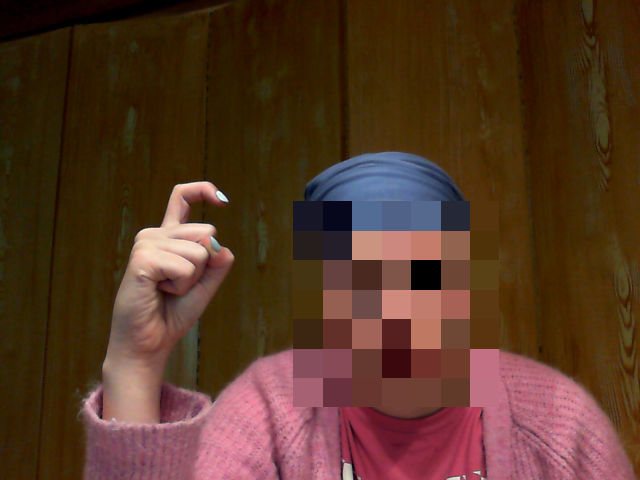}
	\end{subfigure}
	\hfill
    \begin{subfigure}[b]{0.24\textwidth}
		\centering
		\includegraphics[width=\textwidth]{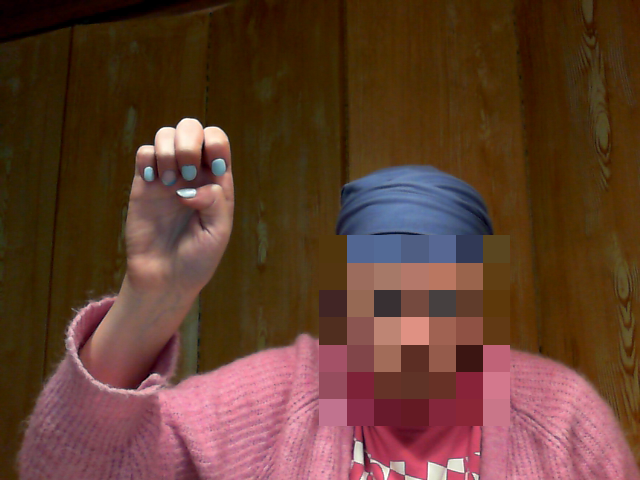}
	\end{subfigure} \\[1pt]
	
	\begin{subfigure}[b]{0.24\textwidth}
		\centering
		\includegraphics[width=\textwidth]{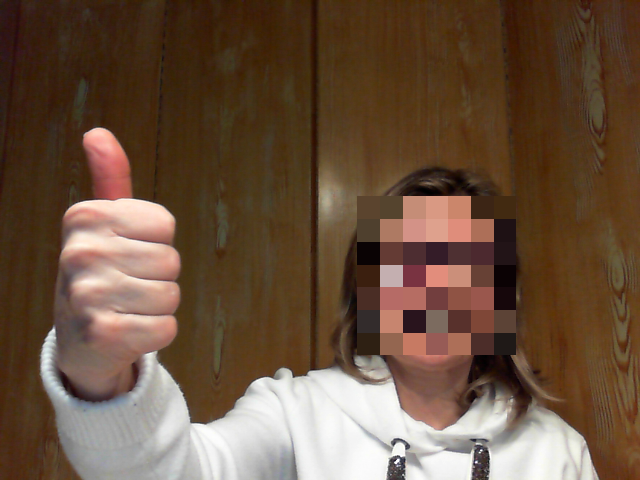}
	\end{subfigure}
	\hfill
    \begin{subfigure}[b]{0.24\textwidth}
		\centering
		\includegraphics[width=\textwidth]{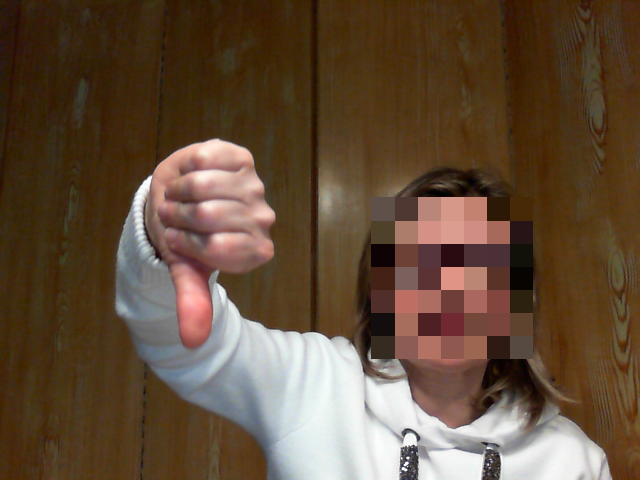}
	\end{subfigure}
	\hfill
	\begin{subfigure}[b]{0.24\textwidth}
		\centering
		\includegraphics[width=\textwidth]{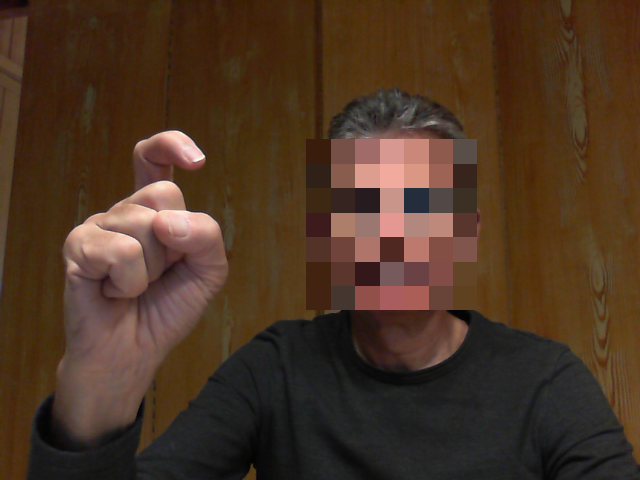}
	\end{subfigure}
	\hfill
    \begin{subfigure}[b]{0.24\textwidth}
		\centering
		\includegraphics[width=\textwidth]{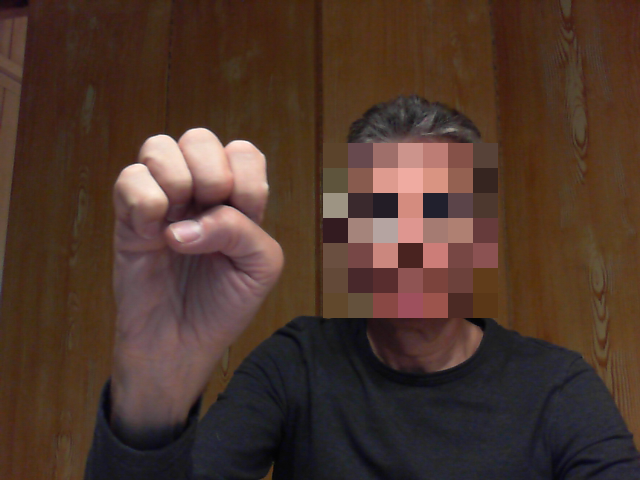}
	\end{subfigure}
	\caption[Example images of the HandGest custom dataset]{Example images of the HandGest custom dataset. Four static gestures performed by four subjects.}
	\label{fig:handgest_examples}
\end{figure}

In addition to the proposed framework, in this paper we present the HAGIL HandGest dataset \footnote{https://xaviercucurull.ml/HAGIL}, a custom hand gesture dataset that was built with the goal of expanding the existing UniPd dataset, by adding new gestures that can be used in an incremental learning scenario. To construct this dataset, some signs from the ASL alphabet were used together with other easy to remember hand gestures. In order to employ different subjects for the train, validation, and test splits, four subjects were used. Figure \ref{fig:handgest_examples} shows some example images from the HandGest dataset. Note that to preserve the privacy of the subjects, their faces were automatically pixelated with the help of a simple Haar cascade \cite{jones2001robust} face detector.

\section{Experiments}\label{sec_experiments}

In this section, first the datasets used for the state-of-the-art experiments are described. Next, the incremental learning experiments, which compare different CL methods and evaluate different parameters, are presented. Finally, results regarding execution times are provided. All the experimental results are averaged over 10 runs, and the order in which new gestures are learned is shuffled at each run

\subsection{Datasets}\label{subsec_dataset}
In order to evaluate the applicability and performance of the HAGIL framework, two group of tests are made. First, the UniPd dataset is extended with the previously presented HAGIL HandGest to compose a dataset consisting of 38 unique symbols. This dataset is referred to as \textit{UniPd+HandGest}. For this combined dataset, the number of training samples is low, ranging from 25 to 50 examples for each unique gesture. Although this amount of data may not be suitable for training a large neural network, good results are obtained using HagilNet. Moreover, this short number of image samples reflects what can be a real use-case scenario, where a robot needs to learn from the interaction with a human user and time and memory constraints limit the number of samples that can be obtained.

Furthermore, two ASL Alphabet datasets are used to evaluate the performance of the proposed framework. The ASL Finger Spelling dataset \cite{pugeault2011spelling} contains 24 static hand gestures that correspond to the static signs of the ASL alphabet. The Kaggle ASL Alphabet dataset \cite{nagaraj2018ASL}  \cite{rasband2018ASL}, which contains 28 static gestures, includes all the ASL alphabet (with a static version of the \textit{j} and \textit{z} letters) and two additional signs.

For all the experiments, different subjects were used for the train, validation, and test splits. This split by subject is really important as to report accurate results close to real-world performance of the model in the wild.

\subsection{Incremental Learning Experiments}\label{subsec_sota}

% In a continual learning setting, $a_{t, k} \subseteq [0, 1]$ designates the accuracy of task \textit{k} after learning task \textit{t}, where $k \le t$ \cite{masana2020class}. In order to compare the overall incremental learning process, the \textit{average accuracy}, defined as $A_{t} = \frac{1}{t}\sum_{i=1}^{t}a_{t, i}$, is used. In addition to that, the average accuracy of classifying the new classes and the previously learned classes by separate is also taken into account, as is done in \cite{lungu2019incremental}.

% In order to compare the results, the regularization strategy LwF will be also trained in an incremental manner. This IL approach does not make use of an exemplar memory, so it will serve as a lower baseline. Besides that, a \textit{Joint} training scenario, where all data is available at each training phase, will be used as an upper bound baseline.

%In this section we perform some experiments to study the impact of some choices of EUREKA on the hand gesture recognition performance.

This section presents the incremental learning tests that were done in order to evaluate the performance of two rehearsal IL approaches: IL2M \cite{belouadah2019il2m} and iCaRL. In addition to that, a regularization approach whicch does not use an exemplar memory (LwF \cite{li2017learning}) is used as a lower baseline and Joint training, in which the model has access to all the training data from previous classes throughout all the IL process, is used as an upper baseline. The IL experiments consist of initially training the model with two classes and then learning the rest of the classes incrementally one by one.

In the following, two main groups of experiments are presented. First, various tests are performed using different numbers of exemplars per class. Then, a second group of experiments evaluates different numbers of epochs used in the incremental tasks. Last, an ablation study is conducted to evaluate the contribution of the different components of the finally selected IL approach. For these experiments, the UniPd+HandGest dataset is used.

\subsubsection{Number of Exemplars}
In these experiments, the performance of using an increasing number of exemplars per class is evaluated, with the maximum number of training epochs set to 50. The goal is to find the lowest rehearsal memory size that leads to a good accuracy, as one of the main goals of continual learning is to avoid storing many data examples from the classes already learned.

\begin{figure}[t]
	\begin{subfigure}[b]{\textwidth}
		\centering
		\includegraphics[width=\textwidth]{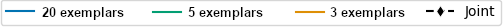}
	\end{subfigure}
	\hfill
    \begin{subfigure}[b]{0.49\textwidth}
		\centering
		\includegraphics[width=\textwidth]{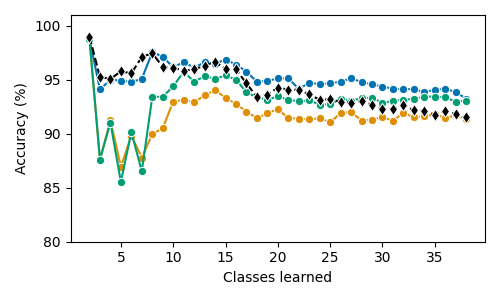}
		\caption{iCaRL}
		\label{subfig:icarl_nexemplars_hagilnet3}
	\end{subfigure}
    \hfill
	\begin{subfigure}[b]{0.49\textwidth}
		\centering
		\includegraphics[width=\textwidth]{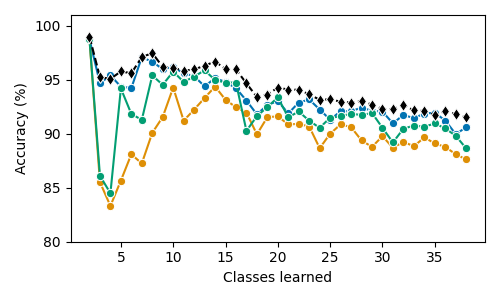}
		\caption{IL2M}
		\label{subfig:il2m_nexemplars_hagilnet3}
	\end{subfigure}
	
    \caption[Experimental results for different number of exemplars per class for iCaRL. Accuracy at each task]{Experimental results for different number of exemplars per class, evaluated on the UniPd+HandGest dataset. Accuracy at each task. For clarity, standard deviation is not plotted.}
    \label{fig:n_exemplars_hagilnet3_icarl_acctask}
\end{figure}

In Figure \ref{fig:n_exemplars_hagilnet3_icarl_acctask}, the average task accuracy for different numbers of exemplars is illustrated for iCaRL and IL2M. In addition, results for the Joint learning scenario are also provided. It can be observed that for the initial tasks, when only a reduced number of classes have been learned, the system lacks stability and exhibits a high variance in the classification accuracy. As more classes are learned, the accuracy results become stable. This instability is higher for a rehearsal memory size of 1 exemplar per class and for IL2M. As iCaRL uses a nearest-exemplar-mean (NEM) classifier, it shows that it can handle better the situation where only a few number of classes are learned with a limited rehearsal memory. It is remarkable to note that iCaRL with 5 exemplars achieves a final accuracy higher than that obtained with Joint training. Therefore, it seems that the NEM classifier of iCaRL provides a robust method to classify the static hand gestures of the UniPd+HandGest dataset.

\subsubsection{Number of Incremental Epochs}

Being able to incrementally train a system with a low number of epochs, while achieving a good accuracy, is something desirable in a real-world scenario, as it reduces training time. For these experiments, the initial 2 classes are trained for 50 epochs. Next, for the rest of the CL episodes, different number of epochs are tested. In this case, a rehearsal memory of 5 exemplars per class selected by herding \cite{welling2009herding} is chosen.

\begin{figure}[!htb]
\centering
    \begin{subfigure}[b]{0.85\textwidth}
		\centering
		\includegraphics[width=\textwidth]{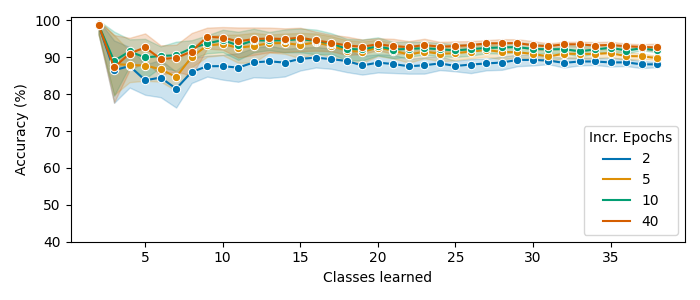}
		\caption{iCaRL}
	\end{subfigure}
    \hfill
	\begin{subfigure}[b]{0.85\textwidth}
		\centering
		\includegraphics[width=\textwidth]{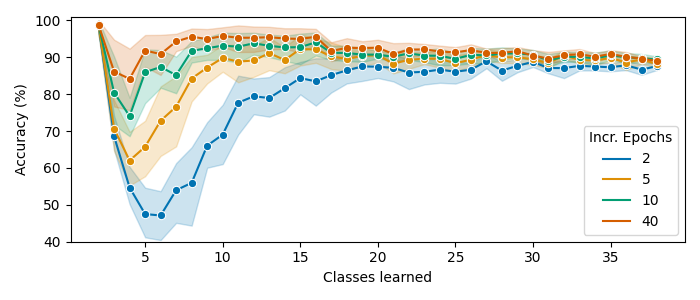}
		\caption{IL2M}
	\end{subfigure}

	\caption[Experimental results for different number of incremental epochs. Accuracy at each task]{Experimental results for different number of incremental epochs. Accuracy at each task, evaluated on the UniPd+HandGest dataset. Colored regions show standard deviation.}
	\label{fig:n_epochs_taskacc}
	
\end{figure}

The results (see Figure \ref{fig:n_epochs_taskacc}) show that iCaRL does not suffer from the cold start problem as much as IL2M, as it obtains good accuracy results from the beginning. In addition, using just 10 incremental epochs gives a performance comparable to using 40 epochs, which can be used to significantly reduce training time.

\subsubsection{Ablation Study}

\begin{table}[!htb]
\centering
    \begin{tabular}{@{}lc@{}}
    \toprule
    \textbf{Method} & \multicolumn{1}{l}{\textbf{Accuracy (\%)}} \\ \midrule
    Joint           & 94.29 ± 4.48                               \\
    HAGIL           & 93.04 ± 5.25                               \\
    iCaRL -kdl      & 92.52 ± 4.85                               \\
    iCaRL -kdl-NEM  & 91.23 ± 5.87                               \\
    iCaRL Raw       & 86.05 ± 7.04                               \\
    LwF             & 16.09 ± 19.59                              \\ \bottomrule
    \end{tabular}
    \caption[HAGIL ablation study experimental results]{HAGIL ablation study experimental results, evaluated on the UniPd+HandGest dataset.}
    \label{tab:ablation_tests}
\end{table}

Table \ref{tab:ablation_tests} provides the average accuracy results for the ablation study, as well as for the upper baseline Joint and the lower baseline LwF. The HAGIL method, which refers to iCaRL with the AllEuclidean+WristDiff+AllDiff feature representation, a rehearsal memory of 5 exemplars per class selected by herding, and 15 incremental epochs, is the IL method that gets the best accuracy. Not using the knowledge distillation loss (\textit{iCaRL -kdl}), slightly degrades the final accuracy, although it obtains good results for the initial learned classes. Removing the NEM classifier (\textit{iCaRL -kdl-NEM}), which is a main component of the iCaRL approach, also impacts the final average classification accuracy, and particularly the accuracy results when only few classes are learned are negatively affected. This effect was also observed in Figure \ref{fig:n_epochs_taskacc} when comparing iCaRL to IL2M. Finally, using the \textit{Raw} feature representation causes the average accuracy to drop. This indicates that selecting a good feature representation is a key aspect of obtaining a good classifier. 

%-------------------------------------------------------------------
\subsubsection{ASL Alphabet Experiments}

In the final experiments, two datasets based on the ASL datasets are evaluated: the Kaggle ASL Alphabet and the ASL Finger Spelling datasets. Although these datasets have less classes than the previously used UniPd+ HandGest dataset, they can be more difficult to learn as the inter-class similarity is higher. For these experiments, HAGIL is used with the configuration found on the previous experiments: iCaRL, 5 exemplars per class with herding, and 15 incremental epochs. The experiment scenario is defined as before, initially learning 2 classes, and learning the rest one by one. It is important to note that the ASL datasets, in contrast to the UniPd and Handgest databases, have hundreds of instances per class, so using a rehearsal memory of 5 exemplars per class represents only a small fraction of the whole training data.

\begin{figure}[t]
\centering
    \begin{subfigure}[b]{0.8\textwidth}
		\centering
		\includegraphics[width=\textwidth]{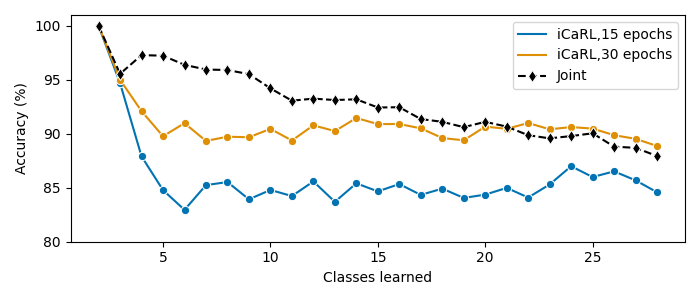}
		\caption{Kaggle ASL Alphabet}
	\end{subfigure}
    \hfill
	\begin{subfigure}[b]{0.8\textwidth}
		\centering
		\includegraphics[width=\textwidth]{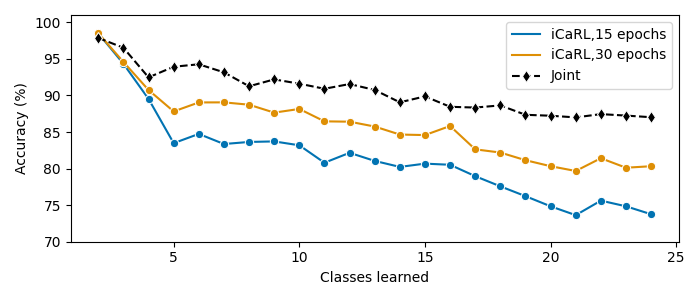}
		\caption{ASL Finger Spelling}
	\end{subfigure}

	\caption[Experimental results for the ASL datasets. Accuracy at each task]{Average accuracy for different number of incremental epochs, using 5 exemplars per class, evaluated on the Kaggle ASL Alphabet and ASL Finger Spelling datasets. For clarity, standard deviation is not plotted.}
	\label{fig:asl_il_experiments}
\end{figure}

As seen in Figure \ref{fig:asl_il_experiments}, HAGIL obtains reasonably good accuracy results in the ASL datasets. However, since the amount of training data available for these datasets is greater than for the previously evaluated UniPd+Handgest dataset, increasing the number of incremental epochs significantly increases the model average accuracy. Comparing the incremental learning with the Joint baseline, we can see that Joint obtains better overall results for both datasets. For the Kaggle ASL Alphabet, however, the IL approach stabilizes after learning the initial 7 classes and maintains a similiar accuracy for the subsequent IL episodes, while the accuracy of the Joint baseliene keeps decreasing. The ASL Finger Spelling dataset, on the other hand, experiences more forgetting and its accuracy performance degrades as more classes are learned. Nevertheless, the obtained results are remarkably good, given the fact that only 5 exemplars are used, compared to the hundreds of training instances. In addition to that, this dataset is more challenging because most of the signs are presented in different rotations and orientations, making it more difficult to learn them with a small rehearsal memory.

\subsection{Execution Times}
An important part of the deployment of an AI system to be used in a real-life scenario is the computation time required by each of the components, as this has a direct impact on the performance of the model in real-time settings. 

As seen in Table \ref{tab:execution_times}, the landmarks extraction is one of the steps which takes more time. Once the landmarks are extracted, the rest of the operations run in a shorter time. HAGIL adds a latency of $34.2 + 2.73 + 0.61 = 37.5ms $, which theoretically would allow an application to run at a maximum rate of around $1/37.5 \times 1000 = 26$ fps.

In order to evaluate the time to train the model, a new experiment is conducted with the Kaggle ASL Alphabet dataset. In this experiment, a model is initially trained with 27 classes, and then it is trained to learn a new class. For both the HAGIL and Joint approaches, the number of epochs used to learn the new symbol is set to 15. As observed in the results, HAGIL takes advantage of the rehearsal memory and obtains a reduced training time of less than 10\% of the time required to learn a new class using all the data from the previously learned classes. 

\begin{table}[t]
\centering
\begin{tabular}{@{}lc@{}}
    \toprule
    \textbf{Operation}                & \multicolumn{1}{l}{\textbf{Execution time}} \\ \midrule
    Mediapipe Landmark Extraction     & 34.2 ms                                     \\
    Feature Representation            & 2.73 ms                                     \\
    Model Inference                   & 608 us                                      \\ \midrule
    Model Training (HAGIL)            & 7.76 s                                      \\
    Model Training (Joint)            & 79.73 s                                     \\ \bottomrule
    \end{tabular}   
    \caption[HAGIL execution times analysis]{Analysis of the execution time of the different components of the HAGIL framework. Training time evaluated by initially training a model with the first 27 classes of the Kaggle ASL Alphabet dataset, and then learning a new class.}
\label{tab:execution_times}
\end{table}

\section{Conclusions}\label{sec_conclusion}
In this work we have presented a system that is capable of incrementally learning static hand gestures from RGB images, by using hand landmarks extraction and the iCaRL class-incremental CL strategy.

To the best of our knowledge, this is the only work that addresses incremental learning of hand gestures from images, and together with the introduced dataset, establishes a benchmark that can benefit both the robotics and continual learning community. HAGIL has proven to be a high performance method that has obtained very competitive results in a variety of datasets.

Moreover, in this work, we have evaluated two popular rehearsal CL techniques (iCaRL and IL2M) in a task other than image classification, as is done in most of the related work. It has been observed that the use of an exemplar memory represents a key component to obtain good accuracy results in an incremental learning scenario. However, it has been possible to keep this memory to a reduced set of exemplars, representing just a small fraction of the original training data, thus reducing the training time and the memory requirements of the system. Thus, the nearest exemplar mean classifier used in iCaRL has proven to be more efficient than the softmax classifier of IL2M when few exemplars and a reduced number of training epochs are used.

Furthermore, we are currently working in the application of this method in the communication between human and robot in agriculture applications, specifically in harvesting and pruning vineyards.

\newpage
\balance
\bibliographystyle{IEEEtran}

\bibliography{IEEEabrv,hri.bib}

\end{document}